\documentclass{article}
\usepackage{spconf,amsmath,graphicx,subfigure}
\usepackage{spconf,amsmath,graphicx,xcolor,array}
\usepackage{booktabs}
\usepackage{threeparttable}
\usepackage{multicol}
\usepackage{multirow}
\usepackage{amsmath}

\usepackage{array}
\usepackage{longtable}
\usepackage{rotating}
\usepackage{multirow}
\usepackage{bbding}

\title{Real-time Anomaly Detection with HMOF Feature}

\name{Huihui Zhu, Bin Liu, Guojun Yin, Yan Lu, Weihai Li, Nenghai Yu\thanks{*Corresponding author: flowice@ustc.edu.cn. This work is supported by the National Natural Science Foundation of China (Grant No. 61371192), the Key Laboratory Foundation of the Chinese Academy of Sciences (CXJJ-17S044) and the Fundamental Research Funds for the Central Universities (WK2100330002, WK3480000005).}}
\address{CAS Key Laboratory of Electromagnetic Space Information\\University of Science and Technology of China, Hefei, China}

\begin{document}
%\ninept
%
\maketitle
\begin{abstract}
%Anomaly detection is a challenging problem in intelligent video surveillance, and  In this paper, we propose a method for real-time anomaly detection. Each video is defined as a set of non-overlapping patches by using KNN Matting foreground detection algorithm, and we describe these patches by our proposed new feature called Histogram of Magnitude Optical Flow(HMOF). This feature is focused on video speed, and captures the motion feature from patches. Then, we use auto-encoder to change the HMOF feature distribution for better clustering. By Gaussian Mixture Model(GMM) Classifiers, we can distinguish normal activities and anomalies in videos. Experimental results show that our algorithm outperforms the state-of-the-art methods, and can reliably detect and localize anomalies in real-time.
Anomaly detection is a challenging problem in intelligent video surveillance. Most existing methods are computation-consuming, which cannot satisfy the real-time requirement. In this paper, we propose a real-time anomaly detection framework with low computational complexity and high efficiency. A new feature, named Histogram of Magnitude Optical Flow (HMOF), is proposed to capture the motion of video patches. Compared with existing feature descriptors, HMOF is more sensitive to motion magnitude and more efficient to distinguish anomaly information. The HMOF features are computed for foreground patches, and are reconstructed by the auto-encoder for better clustering. Then, we use Gaussian Mixture Model (GMM) Classifiers to distinguish anomalies from normal activities in videos. Experimental results show that our framework outperforms state-of-the-art methods, and can reliably detect anomalies in real-time.
\end{abstract}

\section{Introduction}
\label{sec:intro}

%In recent years, terrorist attacks continues to cause social panic, security management for the lack of early warning mechanism in public places which leads to many tragic events. Therefore social intelligence technology has been rapidly developed. However, observation by human and early warning requires a lot of manpower and material resources, thus it is inefficient. Therefore, we need realtime and effective anomaly detection algorithm that can simultaneously deal with multiple video surveillance scenes and detect abnormal behaviours in real-time.
Anomaly detection and localization in intelligent video surveillance is a significant task due to the growing needs of public security.
In real life, the definition of abnormalities in the video is varied. For example, a runner is seen as normal on the track and field, while it will be regarded as abnormal in the square. Therefore, it is difficult for us to use the same standard to measure all the scenes. A video event is usually considered as an anomaly if it is not very likely to occur in the video \cite{cong2011sparse}. Thus we need the normal monitor video of the scene to establish a normal model, which identifies the anomaly in the detection.

\begin{figure*}[tp]
\centering
\includegraphics[height=4.545cm, width=12.8cm]{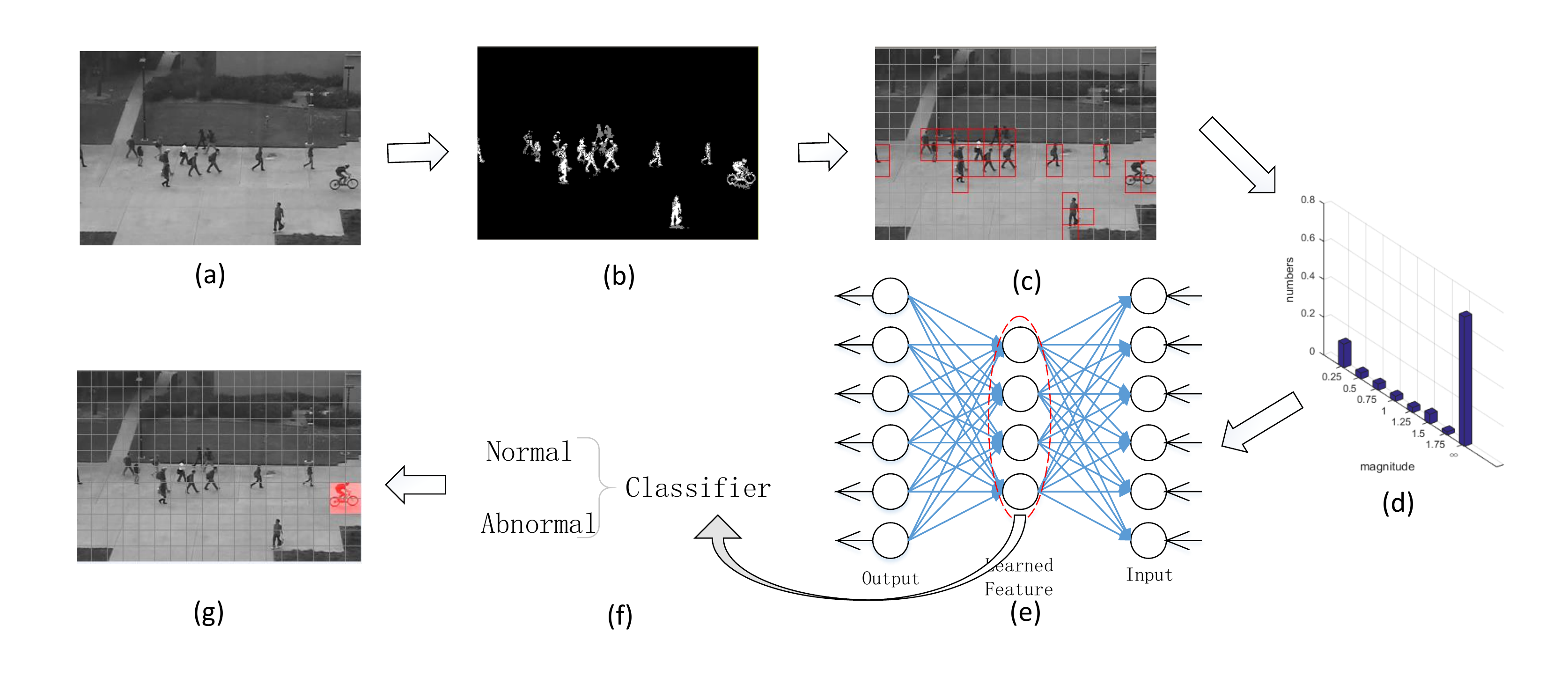}
% \caption{ROC of our work compared to Sabokrou's work in \cite{real-time}}
\caption{Framework of our proposed method. (a)~Input frames. (b)~Extracted foreground. (c)~Foreground patches. (d)~HMOF features. (e)~Auto-Encoder. (f)~GMM Classifier. (g)~Detected anomalies.}
\label{fig:figure1}
\end{figure*}

In recent studies, the sparse representations of events \cite{cong2013video} in videos have been widely explored. The proposed models in \cite{cong2013video,li2014anomaly} achieve favorable performance in global abnormal events (GAE), however they often fail in the local abnormal events (LAE). In order to solve this problem, some methods have proposed trajectory-based anomaly detection methods, such as particle trajectories \cite{wu2010chaotic}, tracking trajectories \cite{jing2013modeling} and so on. Such methods tend to behave very well in simple sparse scenes, but their performance in complex scenes is severely degraded. Some methods take account into the energy distribution characteristics of the crowd. When anomaly exists, some energy can be mutated, such as pedestrian loss model \cite{scovanner2009learning}. Some methods divide frames of video into numbers of patches, and then can detect anomalies in LAE by analyzing the patches \cite{saligrama2012video,reddy2011improved}. In addition, some methods utilize features based on the optical flow, such as Histograms of Oriented Optical Flow (HOF) \cite{xu2014video} and Multi-scale Histogram of Optical Flow (MHOF) \cite{cong2011sparse}. These algorithms are time-consuming and can hardly meet the real-time requirements.

Sabokrou et al \cite{sabokrou2015real} use two feature descriptors to extract global and local features. Roberto et al \cite{leyva2017video} extract the HOF features of the foreground area to build a dictionary to detect anomalies. These algorithms can meet the real-time requirements with high detection speeds. However, compared with state-of-the-art methods, there are still some gaps in detection performance.

%In this paper, we propose a new feature called Histogram of Magnitude Optical Flow(HMOF), which efficient to represent motion information. In order to reduce the processing time, we use the foreground extraction algorithm based on KNN matting to extract the video foreground region, so that only the HMOF feature of the foreground region is extracted by the algorithm. Next, the feature is sent into the auto-encoder network to reconstruct the feature. Finally, the anomaly region is detected by the GMM classifiers. Experiments show that our algorithm performance is better than the state-of-the-art methods, and can be done in real-time.
In this paper, we propose a new feature called Histogram of Magnitude Optical Flow (HMOF), which is more efficient to describe motion information. We use the foreground extraction algorithm to extract the video foreground patches, so that only the foreground patches will be processed, which is more efficient. Next, the features are fed into the auto-encoder network to be reconstructed and then classified by the Gaussian Mixture
Model (GMM) Classifiers. Thus we can distinguish the abnormal patches.

The main contributions of our work are as follows:
(1) We present a new feature named HMOF for anomaly detection. Compared with existing feature descriptors, HMOF is more sensitive to motion magnitude and more efficient to distinguish anomaly information;
(2) We propose an algorithm framework for local anomaly detection, which can be applied in real scenes. Because our algorithm outperforms state-of-the-art methods, and can be done in real-time.

The rest of the paper is organized as follows. The proposed method is introduced in Section 2. Section 3 presents the experimental results, comparisons and analysis on UMN and UCSD datasets. Finally, Section 4 concludes the work.

%\begin{table}
%\caption{Comparison of our work with Sabokrou's work}
%\setlength{\tabcolsep}{3pt}
%\centering
%\begin{tabular}{|c|c|c|c|}\hline
% &Sabokrou's work &ours &ours(Without Patch Selection) \\ \hline
% AUC &0.78   &0.86  &0.80 \\ \hline
%Running Time &0.36s per frame &0.41s per frame &0.30s per frame \\ \hline
%EER &0.23   &0.20   &0.24 \\ \hline
%\end{tabular}
%\label{table}
%\end{table}

\section{THE PROPOSED METHOD}
\label{sec:format}

In this section, we illustrate the proposed algorithm in detail. Firstly, we obtain foregrounds patches with KNN matting. Secondly, HMOF features are extracted on foreground patches. Base on the features, we use the auto-encoder network to get the deep features, which will be fed into the GMM Classifiers. The Framework of our method is shown in Fig.1.

\subsection{Foreground detection}
%\label{ssec:subhead}
We split each frame into a number of non-overlapping patches. In order to reduce the number of processing patches, we use foreground detection algorithm to extract foreground area. As a matter of fact, background patches are eliminated, which can speeds up the test phase.

The problem of foreground segmentation is treated as matting, and the matting model is expressed as follows:
\begin{equation}
\emph{I = aF + (1 - a)B}
\end{equation}
where \emph{I}, \emph{F}, \emph{B} is color, foreground color and background color of a pixel in image respectively, \emph{a} is a parameter of segmentation to indicate the former background weight. Here we use the well-known KNN matting algorithm \cite{chen2013knn} to extract the foreground. The extracted foreground is shown in Fig.1 (b). Then we calculate the foreground value of each patch by adding the intensity of every pixel. If the value exceeds a threshold, we regard it as a foreground patch. The extracted foreground patches are shown in Fig.1 (c).

\subsection{HMOF}
%\label{ssec:subhead}

%As the scale of the target changes over time, the dimension of the corresponding optical flow feature descriptor will change. At the same time, the calculation of the optical flow is sensitive to the background noise, the scale change and the direction of motion. Therefore, it is necessary to find a feature based on optical flow that can characterize time domain action information and not sensitive to scale and direction of motion.
Optical flow is good for describing the motion, and HOF is widely used as a motion descriptor. To extract HOF features, the amplitude weighting statistics of the optical flow is calculated in different directions of the optical flow, and the histogram of the optical flow direction information is obtained. However, the HOF features mainly consider the information of the optical flow direction, with less consideration of the optical flow amplitude information. MHOF \cite{cong2011sparse} is based on the HOF, taking account into the optical flow amplitude information, by setting the relevant threshold information for different amplitude range of different optical flow to carry out statistics.

While MHOF uses the amplitude characteristics of the optical flow, it still mainly considers the direction of the optical flow information, which means that MHOF is less sensitive to the motion magnitude. Furthermore, the amplitude threshold is usually an experience parameter. Generally, abnormal behaviors are more sensitive to the amplitude characteristics rather than directional characteristics of the optical flow, such as running, bicycles, cars, skate, etc. The speed of these behaviors are faster than the speed of normal behaviors. To some extent, the directional characteristics of the optical flow is a kind of interference. To reach a better performance, we propose a new motion feature called HMOF based on the amplitude characteristics of the optical flow, which can detect abnormal objects effectively.

%Different from HOF and MHOF, HMOF focuses on the amplitude characteristics of the optical flow. We divide the amplitude of the optical flow into n intervals, then count the number of optical flow points, and use normalized histogram to keep the scale invariance of the HMOF feature. In addition, the amplitude threshold is not the empirical parameter or the maximum threshold of the region to be measured, but the optical flow threshold of the whole training set is sorted from small to large, and the flow amplitude (threshold) at 94-96\% is selected as HMOF The maximum range of the optical flow is to avoid the sudden increase in the maximum optical flow amplitude in order to avoid the sudden increase in the local optical flow of the video. The last interval is [[(n-1) / n] * threshold, +∞], for accommodate all the optical flow points during the test. Extraction HMOF process shown in Fig.2.

The procedure of HMOF feature extraction is shown in Fig.2. Firstly, we need to calculate the threshold \emph{$\delta$} of HMOF.  We sort the amplitudes of optical flow in normal patches of the whole training set in ascending order. Since there inevitably exists some noise when calculating optical flow, we discard the top \emph{5\%} of the optical flow and set \emph{$\delta$} as the maximum amplitude of the remaining optical flow. Then we divide the amplitude of the optical flow into \emph{n} bins. The range of \emph{i}-th bin is \emph{[(i-1)/n$\times$$\delta$, i/n$\times$$\delta$)}. In order to accommodate all the optical flow during the test phase, the range of the last bin is set to \emph{[(n-1)/n$\times$$\delta$, +$\infty$)}. After that, we use the normalized histogram to keep the scale invariance of the HMOF feature.

Fig.3 shows the feature maps of HOF, MHOF and HMOF, from left to right respectively. Among them, the pedestrian and the tree are normal, and the bicycle and the car are regarded as abnormal. It can be seen that the HMOF is more prominent than the HOF and MHOF, and the characteristic distribution is more obvious. The feature distribution of the normal region is more biased towards the low amplitude side, while the abnormal area characteristic distribution is more biased towards the high side, which is conducive to distinguish between abnormalities. Subsequent experiments show that HMOF features perform better than the HOF and MHOF features.

\begin{figure}

{
    \begin{minipage}{2.7cm}
    \centering
    %\label{fig:subfig:a}  %% label for first subfigure
    \includegraphics[height=4.5cm,width=8.7cm]{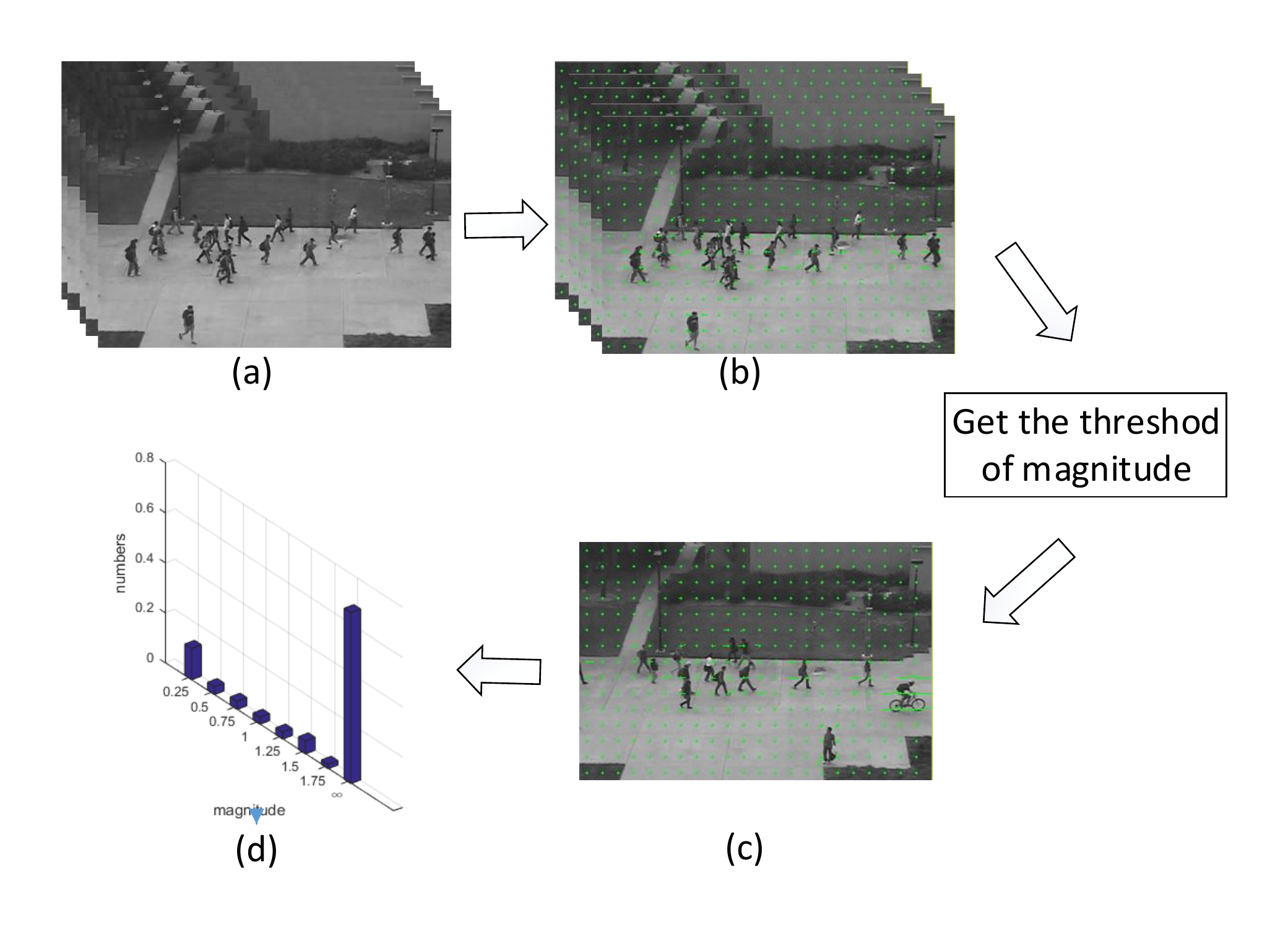}
    \end{minipage}
}
    \caption{The process of extracting HMOF features. (a)~Videos of the training set. (b)~Optical flow of the training set. (c)~An area to be detected. (d)~HMOF of the area.}
%\label{fig:to find the optimal parameters}
\end{figure}

\subsection{Auto-encoder}
%\label{ssec:subhead}

Auto-encoder, consisting of encoder and decoder, is widely used in computer vision area such as video tracking and anomaly detection. Encoder aims to project the input data into the feature space which is constructed by the hidden units, then the input is reconstructed by the decoder. Auto-encoder is usually trained by minimizing the reconstruction error between the input and output. For supervised machine learning, the hidden layer of auto-encoders can be used for feature transformation.

In our case, the space expanded by the hidden units is called the feature space \emph{H}. In the training phase, we use HMOF features of the training set to train the parameters of the auto-encoder. %as the equation implies 公式,
 In the testing phase, we project the HMOF features of input data into feature space \emph{H}. It can be seen that the features distribution of the normal and abnormal samples are quite different in the feature space \emph{H}, which can be easily distinguished by the subsequent classifier.
\begin{figure}[htbp]

{
    \begin{minipage}{8cm}
    \centering
    %\label{fig:subfig:a}  %% label for first subfigure
    \includegraphics[height=5cm,width=6.7cm]{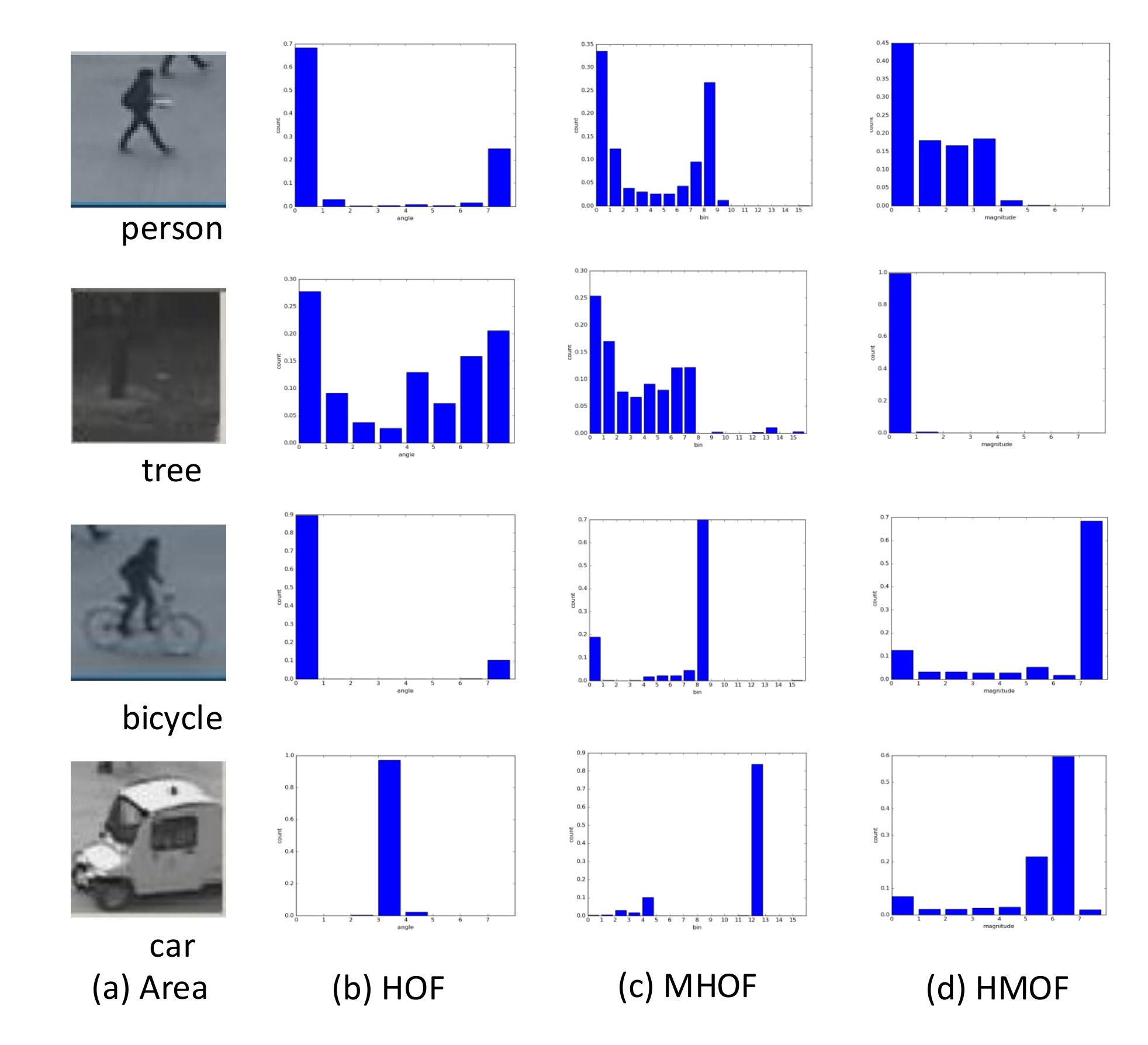}
    \end{minipage}
}
    \caption{the feature maps of HOF, MHOF and HMOF}
%\label{fig:to find the optimal parameters}
\end{figure}
\subsection{Anomaly Classifier}
%\label{ssec:subhead}
The GMM  is a weighed sum of multivariate Gaussian probability densities given by:
%\begin{equation}
%\emph{\rm{P}}(x|\theta ) = \sum\limits_{k = 1}^K {{v_k}}\Phi (x|{\theta _k}{} {)}
%\end{equation}
%where \emph{${v_k}$} is the weight of the \emph{k}th Gaussian model, \emph{$\Phi (x|{\theta _k}{} )$} is the Gaussian density, and \emph{$\theta$} is the collection of all parameters for the GMM.
\begin{equation}
P(\textbf{x}|\Theta ) = \sum\limits_{k = 1}^K {{\lambda _k}{\cal N}(\textbf{x}|{\boldsymbol{\mu}_k},{\boldsymbol{\Sigma}_k})}
\end{equation}
where \emph{$\Theta  = \{ {\lambda _1}, \cdots ,{\lambda _K},{\boldsymbol{\mu} _1}, \cdots ,{\boldsymbol{\mu} _K},{\boldsymbol{\Sigma} _1}, \cdots ,{\boldsymbol{\Sigma} _k}\}$} is the parameter of GMM. \emph{$K$} denote the number of Gaussian components and \emph{${\lambda _k}$} is the weight of the $k$-th Gaussian model. ${\boldsymbol{\mu} _k}$ and \textbf{${\boldsymbol{\Sigma} _k}$} are the mean and covariance matrix respectively. ${\cal N}$($\cdot$) denotes the multivariate Gaussian distribution. The parameters can be estimated by using the maximum-likelihood (ML) estimation. With the GMM method, we can adaptively adjust the decision surface for classification, which can better distinguish anomalies from normal activities in videos.

At first we use all features of the training set to train the GMM Classifiers, then the trained classifier is used to test the features of the testing set. Each feature will get a score after passing the classifier. If it is below a threshold \emph{$\alpha$}, it will be judged as an abnormal:
\begin{equation}
{Patch}{(\textbf{x})  =  }\left\{ \begin{array}{l}
Normal\hspace{0.7cm}score{\rm{(\textbf{x}) > \alpha}}\\
Abnormal\hspace{0.4cm}otherwise
\end{array} \right.\
\end{equation}
where $\textbf{x}$ is the feature fed into the classifiers, and $score(\cdot)$ is the score given by the GMM Classifiers.

In some surveillance scenes, a target may be the formation of several patches due to the limitation of patch-based method. Thus in order to enhance the robustness, we judge a frame as abnormal if the number of abnormal patches exceeds a threshold \emph{$\beta$}. If the number of abnormal patches is below the \emph{$\beta$}, these patches are more likely to be misjudged and we will drop them out from the abnormal candidates:
\begin{equation}
{Frame}{(i)  =  }\left\{ \begin{array}{l}
Normal\hspace{0.7cm}sum{\rm{(p) < \beta}}\\
Abnormal\hspace{0.4cm}otherwise
\end{array} \right.\
\end{equation}
where $\rm{p}$ are the abnormal patches, $sum(\cdot)$ is the number of the abnormal patches in the $i$-th frame.

\label{sec:pagestyle}
\begin{table*}[htbp]
%\caption{Details of run-time (second/frame)}
\centering
{\bf Table 3:} Details of running-time (second/frame) on UCSD Ped2
\begin{tabular}{lcccccc}
\hline
Time (spf)&Foreground&Optical Flow&HMOF&Auto-Encoder&GMM&Total\\ \hline \hline
Ours method&0.011&0.025&0.004&0.006&0.002&0.048\\ \hline
%ours&{\color{red} 0.801}&{\color{red} 0.747}\\ \hline
\end{tabular}

\end{table*}

\section{EXPERIMENTS}
We use two measures to evaluate the results: the frame-level and the pix-level. For the frame-level, if one pixel is detected as an anomaly, the whole frame is considered as an anomaly. For the pixel-level, a frame is deemed to be correctly classified if at least 40\% of the pixels are correctly classified \cite{mahadevan2010anomaly}. We test our algorithm on two datasets: the UMN dataset for GAE detection and the UCSD dataset for LAE detection. The details are shown below.

\subsection{Detection of GAE on UMN dataset}
%\label{ssec:subhead}
The UMN dataset has three different scenes with a resolution of 320 $\times$ 240. In each scene, a group of people are walking in an area, and suddenly all people run away, which is considered to be abnormal. This dataset has no pixel-level ground truth, so we use Area Under the Curve (AUC) of the frame-level to evaluate our method.

We set the patch at a size of 20 $\times$ 20, and the amplitude of the optical flow is divided into 8 bins. The threshold \emph{$\delta$} of HMOF is calculated as 1.04 and the \emph{$\beta$} used to adjudge abnormal frame is set to 3. Some image results are shown in Fig.4. We compare our method with SR \cite{cong2011sparse}, Zh \cite{liu2014abnormal}, MIP \cite{du2013abnormal}, Scan \cite{hu2013unsupervised}. Results are shown in Table 2, which demonstrate that our method outperforms state-of-the-art methods.
\begin{table}

%\caption{Comparison of AUC on the UMN dataset}
%\centering  % 表居中
\centering
{\bf Table 1:} Comparison of AUC on the UMN dataset

\small
\begin{tabular}{lccccc}  % {lccc} 表示各列元素对齐方式，left-l,right-r,center-c
\hline
Scene &SR\cite{cong2011sparse}    &Zh\cite{liu2014abnormal}    &MIP\cite{du2013abnormal}     &Scan\cite{hu2013unsupervised}      &Ours \\ \hline  % \hline 在此行下面画一横线
  1    &99.5\%                       &99.3\%                          &99.6\%                       &99.1\%                        &{\bf 99.8\%}\\  \hline        % \\ 表示重新开始一行
  2    &97.5\%                       &96.9\%                         &94.4\%                         &95.1\%                       &{\bf 98.6\%}\\  \hline        % \\ 表示重新开始一行
  3    &96.4\%                      &98.8\%                         &90.8\%                          &99\%                         &{\bf 99.2\%}\\  \hline        % \\ 表示重新开始一行

\end{tabular}
\end{table}

%\begin{table}
%
%%
%%\centering  % 表居中
%\centering
%%{\bf Table 1:} Comparison of AUC on the UMN dataset
%\Large
%\begin{tabular}
%{l|cccc}  % {lccc} 表示各列元素对齐方式，left-l,right-r,center-c
%\hline
%
%& \multicolumn{4}{c}{ detection from  layer}\\
%mAP                      &5                                   &6                           &7                            &4 \\ \hline  % \hline 在此行下面画一横线
%65.5                      &                        &                      &                        &\Checkmark \\        % \\ 表示重新开始一行
%68.3                      &                        &                        &\Checkmark                       &\Checkmark \\        % \\ 表示重新开始一行
%72.5                      &                                 &\Checkmark                         &\Checkmark                       &\Checkmark \\        % \\ 表示重新开始一行
%74.2                    &\Checkmark                         &\Checkmark                           &\Checkmark                         &\Checkmark \\  \hline        % \\ 表示重新开始一行
%\end{tabular}
%%\vspace{0.5cm}
%\caption{Combining features from different layers.}
%\end{table}

\begin{figure}
\begin{minipage}{3cm}
\centerline{\includegraphics[height=1.89cm,width=2.7cm]{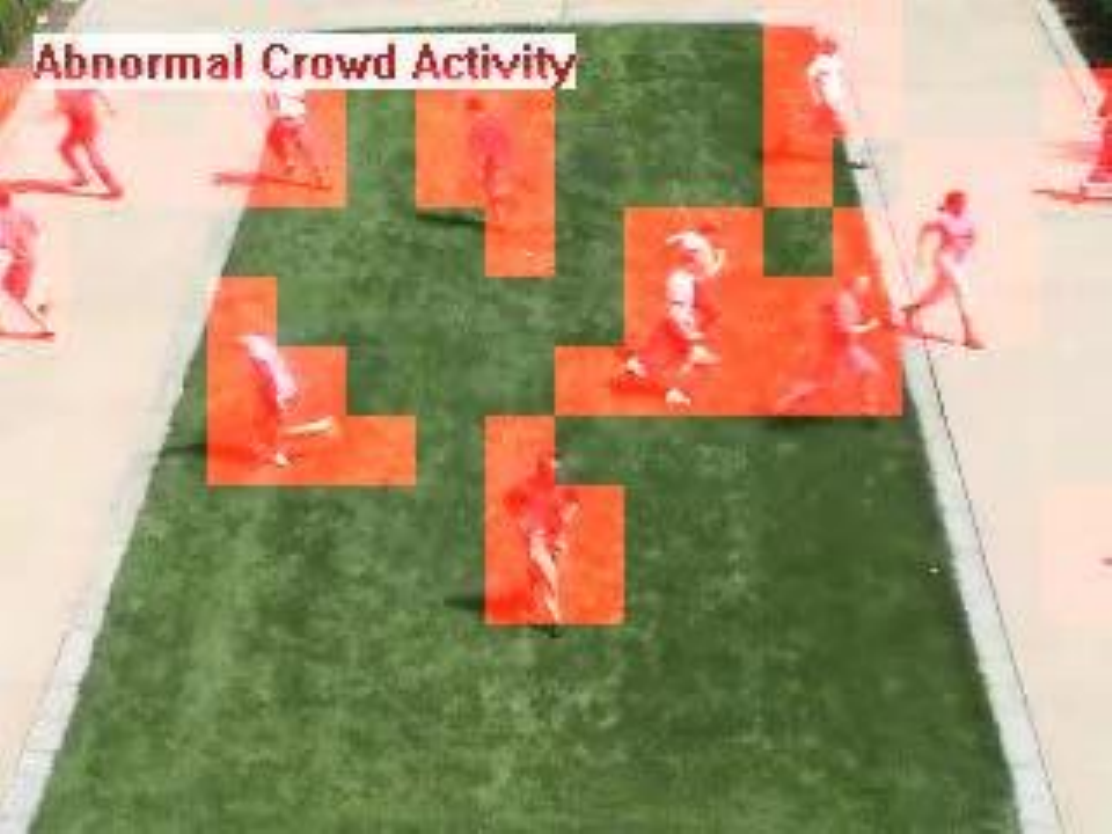}}
%\centerline{}
\end{minipage}
\qquad
\begin{minipage}{1cm}
\centerline{\includegraphics[height=1.89cm,width=2.7cm]{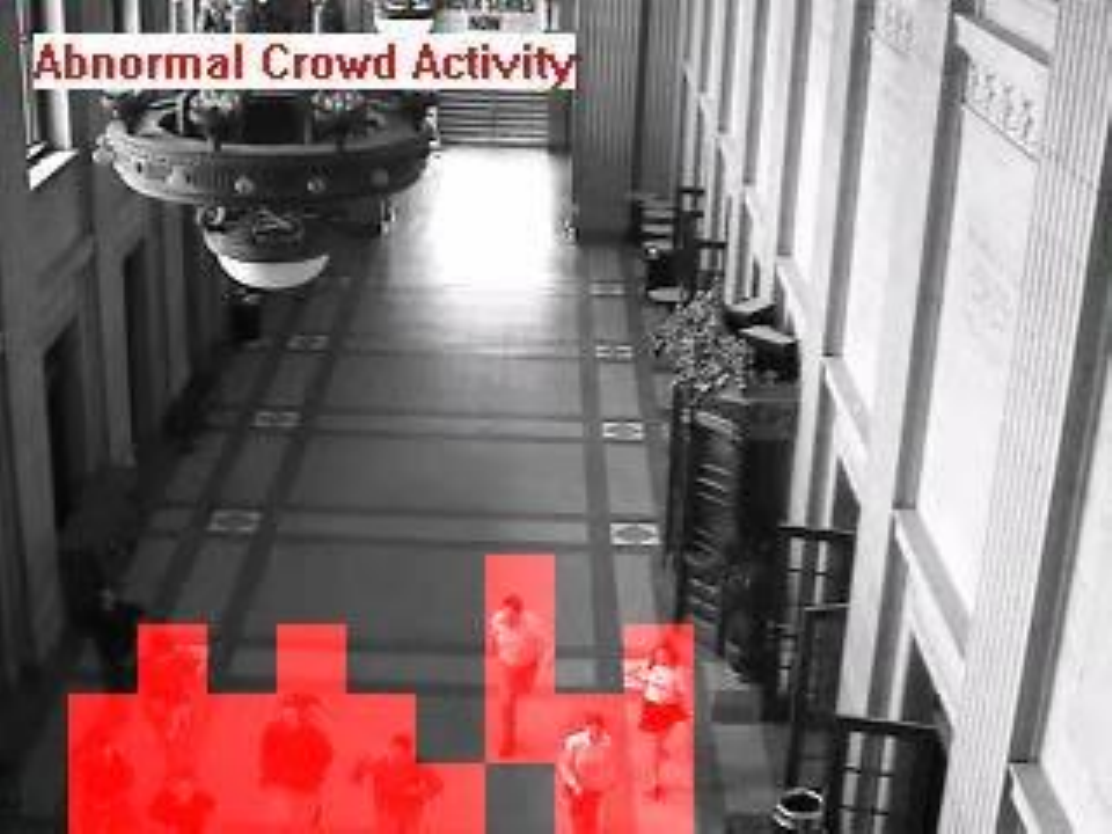}}
\end{minipage}
\qquad
\begin{minipage}{3cm}
\centerline{\includegraphics[height=1.89cm,width=2.7cm]{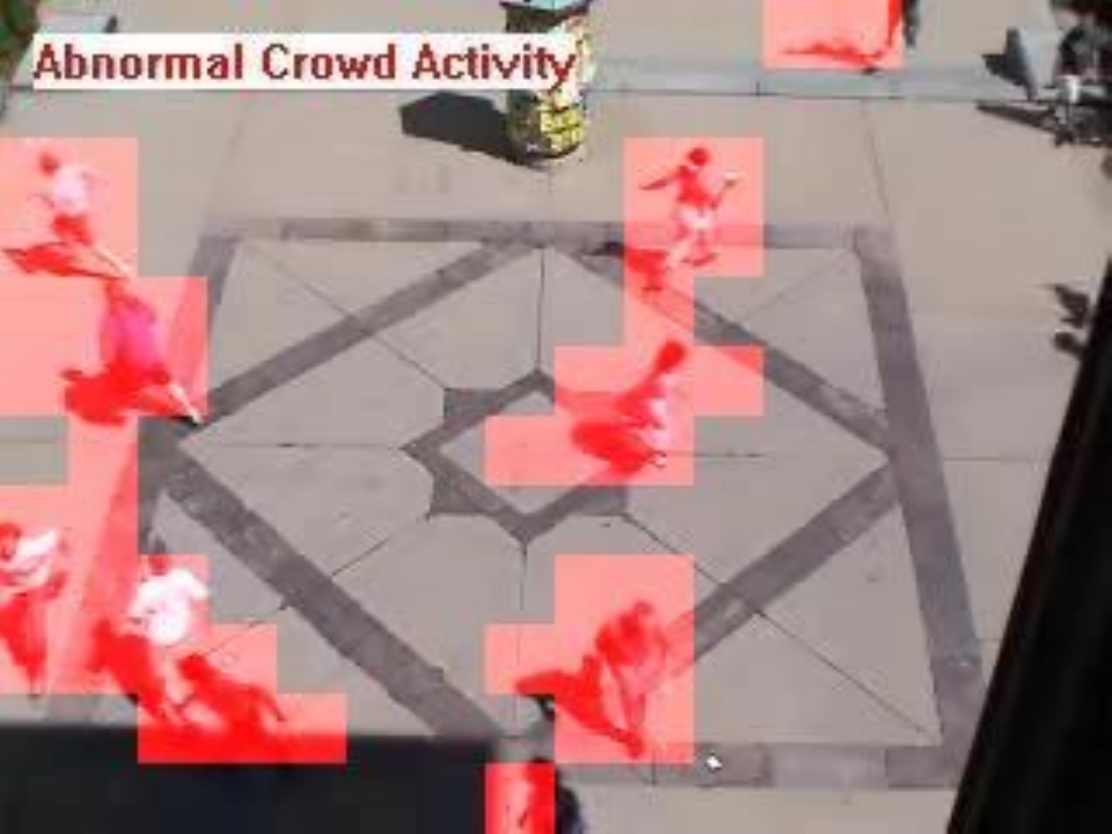}}
\end{minipage}
\\
\begin{minipage}{3cm}
\centerline{\includegraphics[height=1.89cm,width=2.7cm]{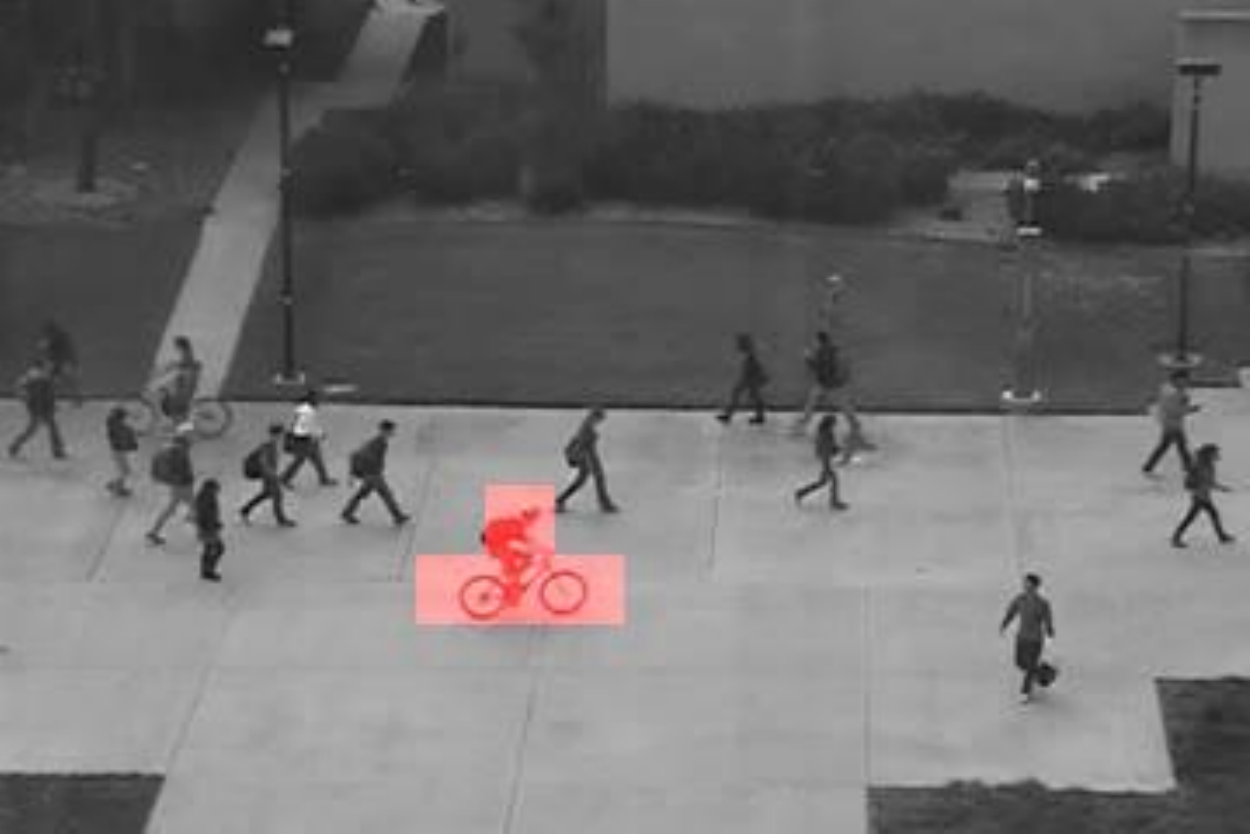}}
\end{minipage}
\qquad
\begin{minipage}{1cm}
\centerline{\includegraphics[height=1.89cm,width=2.7cm]{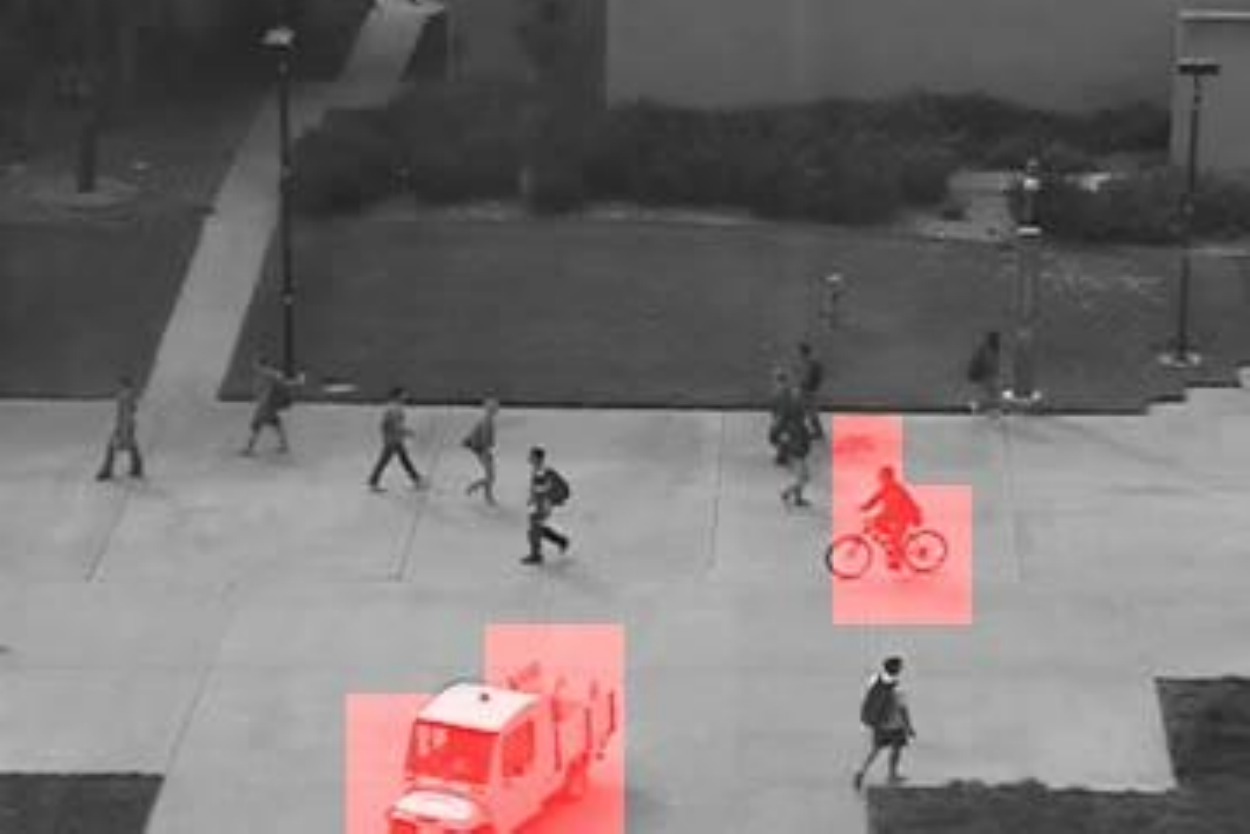}}
\end{minipage}
\qquad
\begin{minipage}{3cm}
\centerline{\includegraphics[height=1.89cm,width=2.7cm]{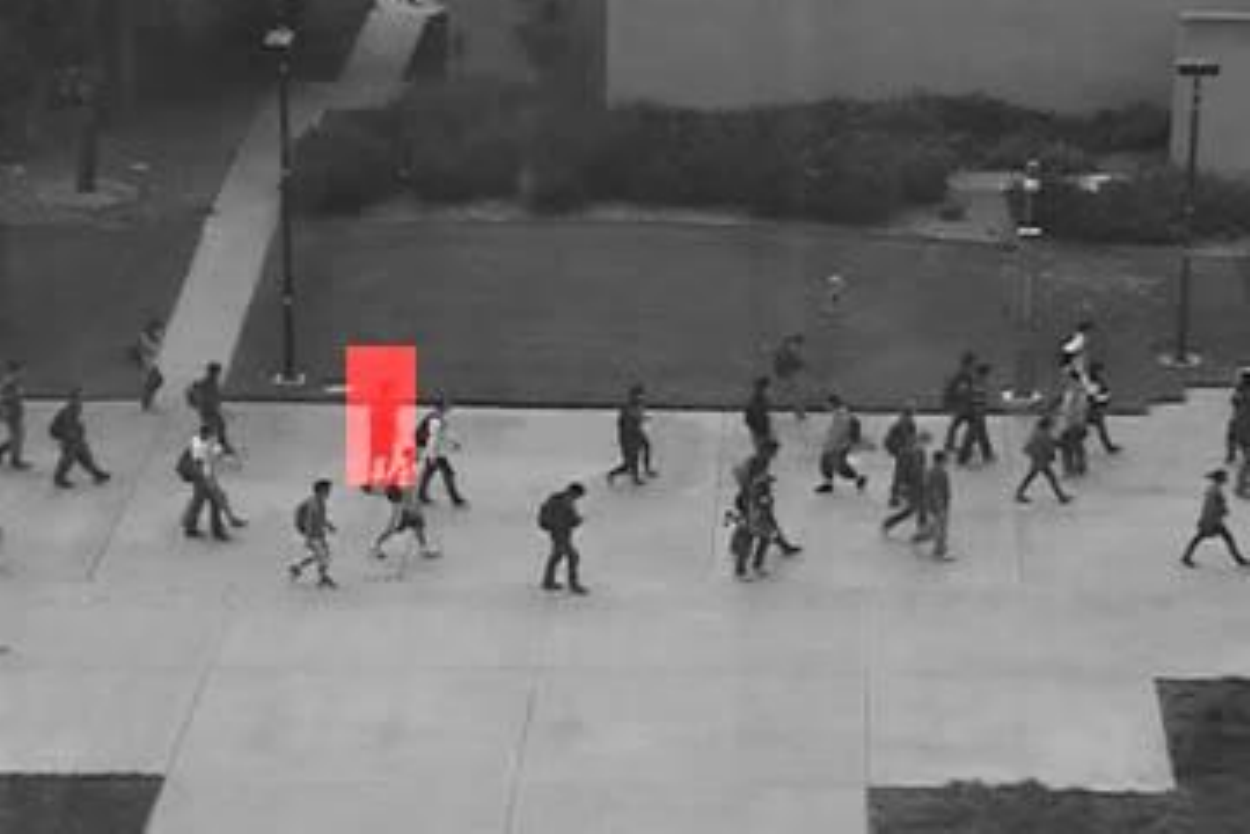}}
\end{minipage}
\caption{Examples of anomaly detection on UMN (top row) and UCSD Ped2 (bottom row)}
\end{figure}

\subsection{Detection of LAE on UCSD Ped2 dataset}
%\label{ssec:subhead}
The UCSD Ped2 dataset has 16 training and 12 testing video clips, and the number of frames of each clip varies. The videos consist of walking pedestrians paralleling to the camera plane, which are recorded with a static camera at 10 fps.

We set the patch at a size of 20 $\times$ 20, and the amplitude of the optical flow is divided into 8 bins. The threshold \emph{$\delta$} of HMOF is calculated as 2.4 and the \emph{$\beta$} is set to 3. Some image results are shown in Fig.4. Our algorithm can detect bikers, cars, skaters, etc. Fig.5 shows the frame-level and pixel-level Receiver Operating Characteristic (ROC) of the UCSD Ped2. Equal Error Rate (EER) for the frame-level and the pixel-level comparisons is shown in Table 2. From Table 2, we can see that if the HMOF features are replaced by HOF or MHOF in the proposed method, the performance is much worse, which indicates the effectiveness of the HMOF features. We can also see from Table 2, compared with other state-of-the-art methods, the proposed algorithm achieves the best performance, with the frame-level EER decreased from 8.2\% to 7.2\%, and the pixel-level EER decreased from 17\% to 14.8\%.

\begin{figure}
\begin{minipage}{4.1cm}
\centerline{\includegraphics[height=4cm,width=4.6cm]{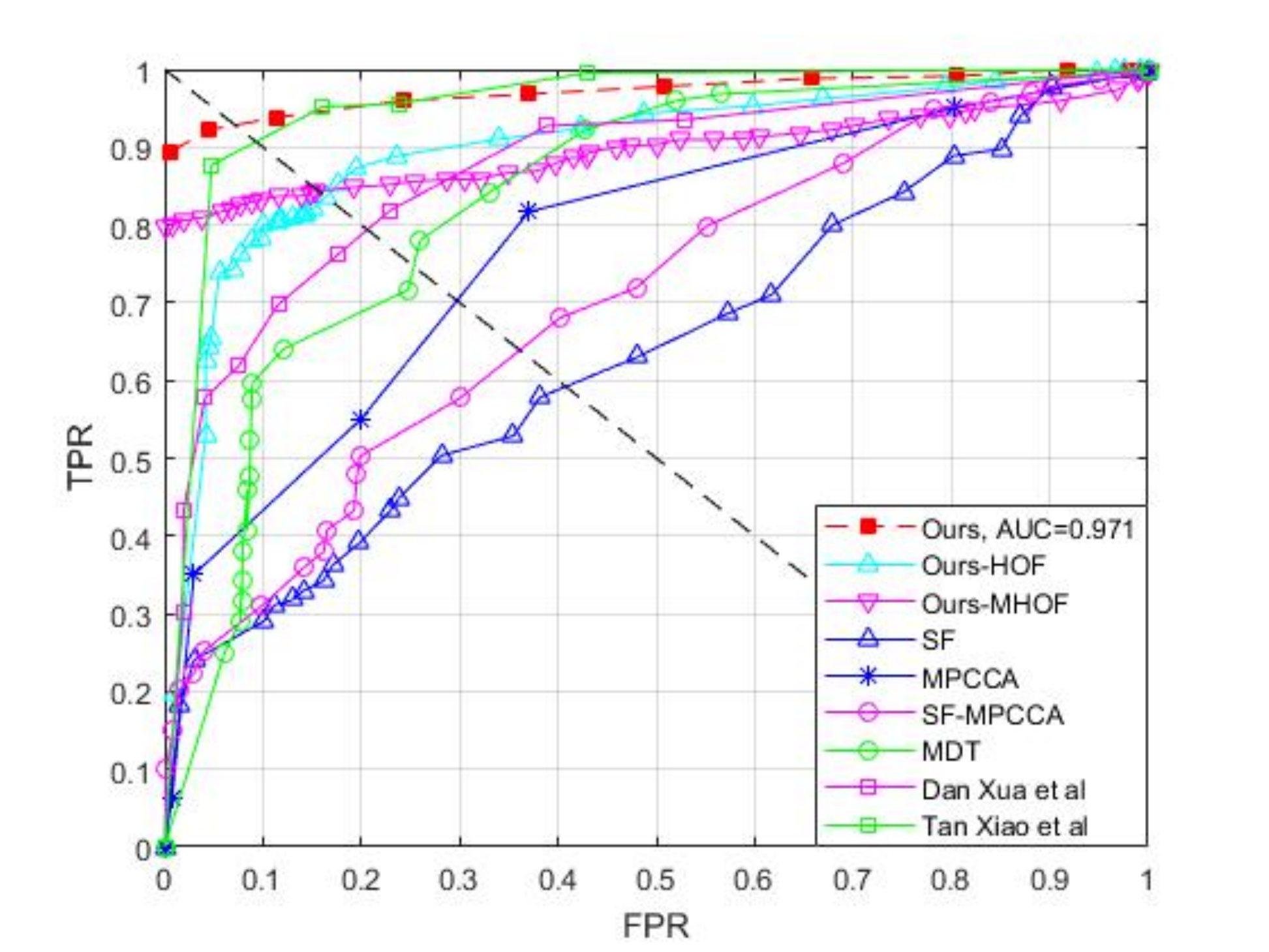}}
%\centerline{}
\end{minipage}
\qquad
\begin{minipage}{3.4cm}
\centerline{\includegraphics[height=4cm,width=4.6cm]{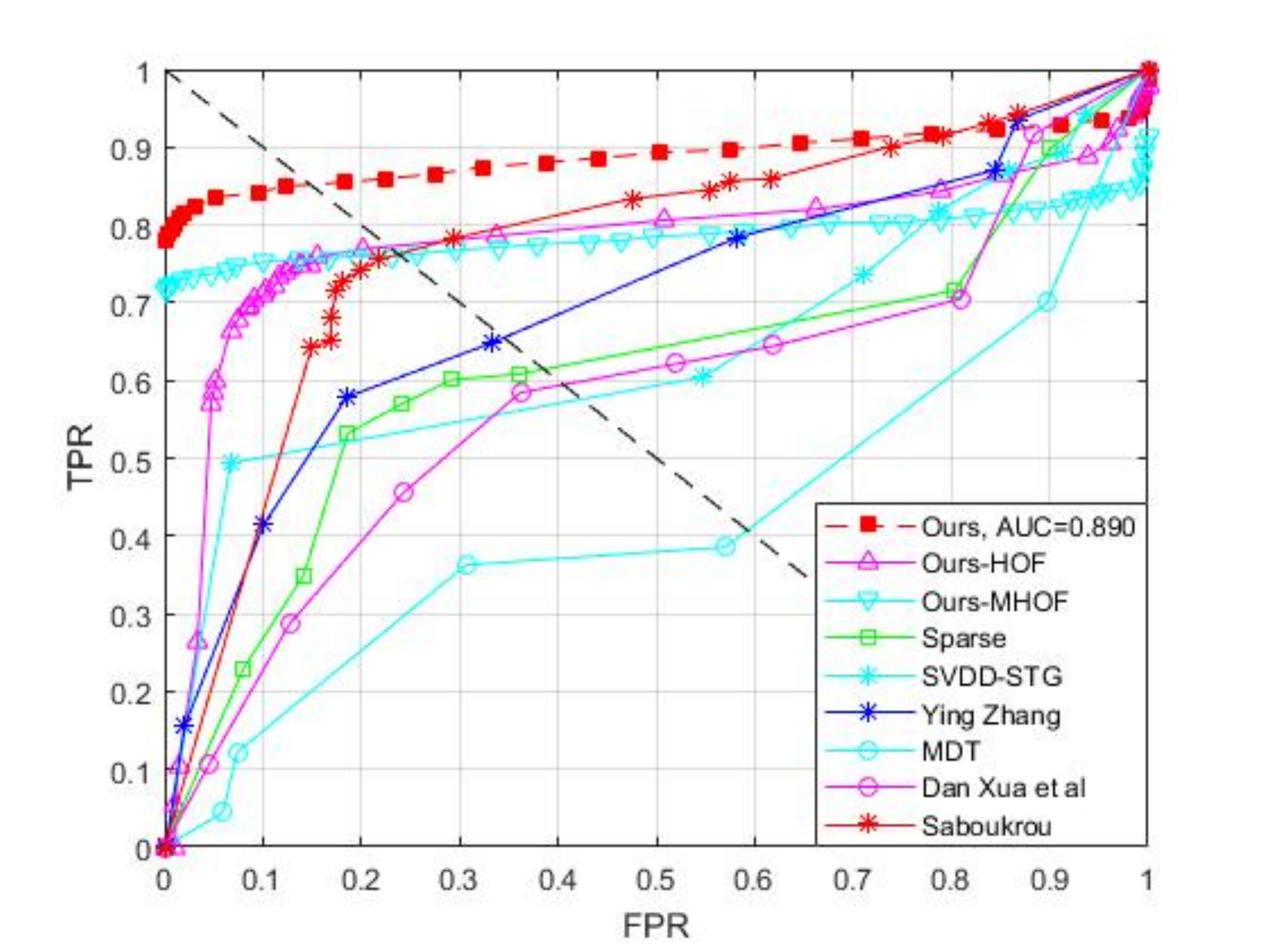}}
\end{minipage}
\caption{ROC comparison with state-of-the-art methods. Left: Frame-level. Right: Pixel-level}
\end{figure}

\begin{table}

{\bf Table 2:} EER for frame-level (FL) and pixel-level (PL) comparisons on UCSD Ped2; we only list first author in this table)% for reasons of available space)

%\caption{EER for frame level and pixel level comparisons on Ped2}
\footnotesize

%\centering  % 表居中
\centering
\begin{tabular}{lcc|lcc}  % {lccc} 表示各列元素对齐方式，left-l,right-r,center-c
\hline
Method                           &FL    &PL              & Method   &FL     &PL   \\ \hline\hline  % \hline 在此行下面画一横线
MDT \cite{mahadevan2010anomaly}       &24\%     &54\%         &Saligrama \cite{saligrama2012video}    &18\%     &-        \\         % \\ 表示重新开始一行
Reddy \cite{reddy2011improved}    &21\%    &31\%            &Ying Zhang \cite{zhang2016combining}         &22\%    &33\%       \\         % \\ 表示重新开始一行
Dan  \cite{xu2014video}             &20\%     &42\%             &Sabokrou \cite{sabokrou2015real}    &19\%    &24\%      \\         % \\ 表示重新开始一行
Rosh \cite{javan2013online}    &17\%     &30\%              &DeepCascade \cite{sabokrou2017deep}      &8.2\%     &19\%       \\         % \\ 表示重新开始一行
Li \cite{li2014anomaly}         &18.5\%    &29.9\%          &Tan Xiao \cite{xiao2015learning}    &10\%     &17\%                 \\         % \\ 表示重新开始一行
IBC \cite{boiman2007detecting}    &13\%    &26\%            &Ours-MHOF     &15.5\%     &23.9\%                 \\         % \\ 表示重新开始一行
Ours-HOF    &16.4\%     &22.8\%              &Ours      &{\bf 7.2\%}    &{\bf 14.8\%}    \\         % \\ 表示重新开始一行
\hline
\end{tabular}
\end{table}

\subsection{Running-Time Analysis}
%\label{ssec:subhead}

%Our method have a low computational complexity. We use a PC with 3.0 GHz CPU to run our method, running-time details on UCSD Ped2 for processing a single frame is provided in Table 1. The total time for detecting an anomaly in a frame is 0.052sec, which proposed that our method can be run in real-time.

The experiments are conducted on a regular PC with Intel-i7-7700 CPU (3.6 GHz) and 8 GB RAM, and the running-time of processing a single frame of UCSD Ped2 is provided in Table 3. Our method is computational efficient with the total time for detecting an anomaly in a frame being 0.048sec, which indicates that our method can be proceed in real-time.

\section{CONCLUSION}
\label{sec:typestyle}

In this paper, we present an anomaly detection method. A new feature named HMOF is proposed. Compared with other feature descriptors, HMOF is more sensitive to motion magnitude, and efficient to represent anomaly information. In our method, HMOF is computed for each foreground area, and is reconstructed by the auto-encoder. Then we use GMM Classifiers to distinguish anomalies from normal activities in videos. Experimental results show that our algorithm outperforms state-of-the-art methods, and can reliably detect anomalies in real-time. Therefore, it can be widely used in real-time surveillance applications.

% \vfill\pagebreak

\bibliographystyle{IEEEbib}
\bibliography{refs}
\end{document}